\documentclass[10pt]{article}
\usepackage[letterpaper]{geometry}
\usepackage{hicss}
\usepackage{times}
\usepackage[none]{hyphenat}
\usepackage{url}
\usepackage{latexsym}
\usepackage{minted}
\usepackage{indentfirst}
\usepackage{graphicx}
\graphicspath{{images/}}
\usepackage[
    style=apa,
  ]{biblatex}
\usepackage{amsmath}
\usepackage{amssymb}
\usepackage{tikz}
\usepackage{dsfont} 
\usepackage{subcaption}

\usepackage[scaled=0.92]{helvet}         
\DeclareCaptionFont{helv}{\fontfamily{phv}\selectfont}

\usepackage{adjustbox}

\newcommand{\ourPara}[1]{\vspace{0.3\baselineskip}\textbf{#1}}

\title{
  MoCap2Radar:  A Spatiotemporal Transformer \\[0.4\baselineskip]
  for Synthesizing Micro‑Doppler Radar Signatures from Motion Capture
}

\author{
  Kevin Chen\\
  Computer Science \& Engineering \\
  The Ohio State University \\
  {\underline{chen.11020@osu.edu}} \\
  \And
  Kenneth W. Parker \\
  Unaffiliated \\ 
  {\underline{kenneth@parkertong.net}} \\
  \And
  Anish Arora\\
  Computer Science \& Engineering \\
  The Ohio State University \\
  {\underline{arora.9@osu.edu}} \\
 }

\date{}

\begin{document}
  \maketitle
  
\begin{abstract}
  We present a pure machine learning process for synthesizing radar spectrograms from Motion‑Capture (MoCap) data. 
  We formulate MoCap‑to‑spectrogram translation as a windowed sequence‑to‑sequence task using a transformer‑based model that jointly captures spatial relations among MoCap markers and temporal dynamics across frames. 
  Real‑world experiments show that the proposed approach produces visually and quantitatively plausible doppler radar spectrograms and achieves good generalizability.
  Ablation experiments show that the learned model includes both the ability to convert multi-part motion into doppler and an understanding of the spatial relation between different parts of the human body.
  
  The result is an interesting example of using transformers for time-series signal processing.
  It is especially applicable to edge computing and Internet of Things (IoT) radars.
  It also suggests the ability to augment scarce radar datasets using more abundant MoCap data for training higher-level applications.
  Finally, it requires far less computation than physics-based methods for generating radar data.
\end{abstract}
\vspace*{1.2\baselineskip}
\noindent
\ourPara{Keywords:} 
cross-modal regression, doppler spectrogram synthesis, machine learning (ML), motion capture (MoCap), spatiotemporal transformer.
\section{Introduction}


Micro-doppler radar data plays a critical role in numerous machine learning applications, such as human activity recognition \parencite{Shi2018, Tan2024}, health monitoring \parencite{Xu2022}, and gesture recognition \parencite{Zhang2017}. This is largely due to several inherent advantages of doppler radar:  it operates under varying lighting and weather conditions, preserves privacy by not capturing visual imagery, is capable of detecting micro-motions through some types of walls, and can be implemented with low-power and limited computing resources. These properties make doppler radar a robust and non-intrusive sensing modality, well-suited for edge computing and IoT devices \parencite{roy2021one, sturzinger2025using}.
	
However, for edge inference and learning tasks, datasets spanning a wide range of activity patterns are not available, and collecting large datasets would require infrastructure that is expensive and not readily available.  These challenges underscore the need for alternative approaches, particularly the synthesis of realistic doppler radar data.

\begin{figure}[t!]
    \centering
    \includegraphics[width=\columnwidth]{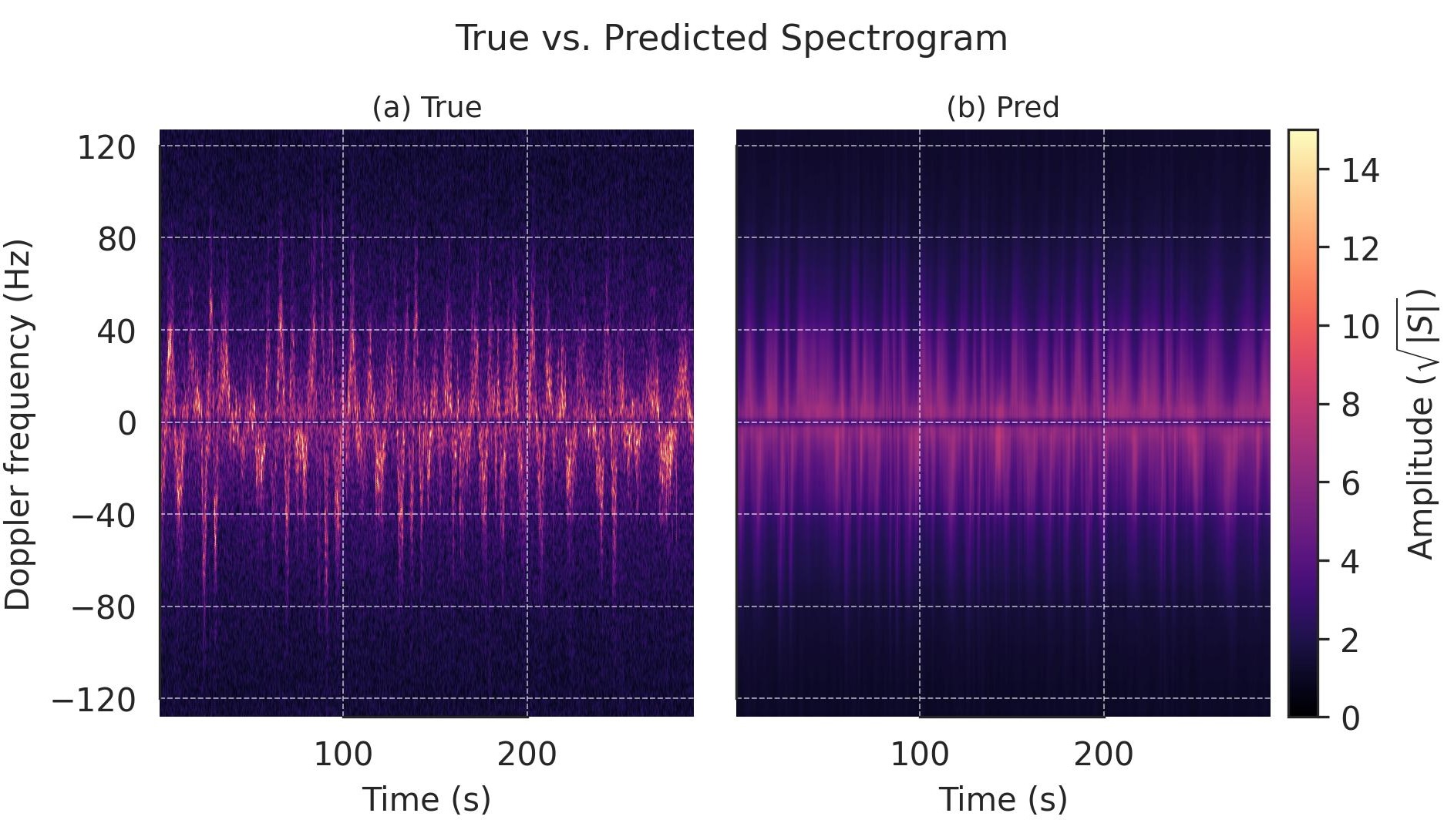}
    \caption{Our MoCap2Radar trained only on straight walks still generates realistic spectrogram for a random walk: left shows radar ground truth, right the model’s prediction from MoCap.}
    \label{fig:money}
\end{figure}

Motivated by the recent work, which we will discuss in the next section, we explore an alternative approach to doppler radar data synthesis: mapping of MoCap sequences to doppler radar spectrograms.
We formulate this translation task as a sequence‑to‑sequence (Seq2Seq) regression problem because both the input MoCap stream and the target radar spectrogram constitute temporally ordered sequences. 
Transformers are particularly well suited to Seq2Seq problems because their global self‑attention models accommodate all source‑target token-interactions in parallel, yielding markedly better long‑range dependency capture and convergence speed than recurrence‑based (LSTM/GRU) or convolution‑based encoder–decoders.

The main contributions of this paper are:%
\begin{enumerate}
    \item We cast doppler–radar data synthesis as a \emph{pure} Seq2Seq machine‑learning problem, dispensing with traditional physics-based simulated solutions.
    \item This work demonstrates, for the first time, the utility of \textit{spatiotemporal transformer} architecture for the task of MoCap-to-radar signal synthesis translation, notably, including single range-bin doppler data.
    \item We provide a complete process for augmenting limited radar data, given MoCap data and a sufficient amount of radar-MoCap paired data.
\end{enumerate}

Figure~\ref{fig:money} shows the result of our \emph{MoCap2Radar} model for human walks, when trained only on straight walks and tested on a randomly meandering walk. The result illustrates a degree of generalization.

The rest of the paper is organized as follows. Section~\ref{sec:related} describes the recent and related work on radar synthesis. Section~\ref{sec:problem} introduces the problem statement and presents our spatiotemporal transformer model to solve the problem.
The data set and its collection are described in Section~\ref{sec:data}. Section~\ref{sec:validation} presents the performance evaluation and ablation studies.
Finally, Section~\ref{sec:discussion} speculates on the implications of our solution.



\section{Related Work}
\label{sec:related}

A common approach for radar data synthesis is to use motion capture (MoCap) or video, to generate corresponding doppler radar signals. 
Compared to radar data, these modalities are more readily accessible and enjoy higher labeling accuracy, making them suitable for synthesizing radar signals. For example, \textcite{Ahuja2021} synthesize doppler radar spectrograms from videos by: fitting video data to 3D human models to estimate poses, computing radial velocities and radar cross-sections for each vertex, and refining the resulting doppler-time plots through a U-Net-based encoder-decoder model. \textcite{Vishwakarma2022} synthesize radar micro-doppler signatures by integrating MoCap animation data with IEEE 802.11 WiFi signal models and employing primitive-shape electromagnetic scattering to generate realistic radar returns across various radar configurations.

These approaches fit the MoCap data to a learned model of humans and then employ physics‑based simulation of the returns from these models. 
To the best of our knowledge, there is no  purely data‑driven, end‑to‑end learning framework that maps visual motion measurements directly to radar spectrograms without intermediate electromagnetic simulation. 
Direct computation, without physics-based simulation, should be more than an order of magnitude more efficient.

Recent advances in human motion modeling have shown that transformer-based architectures are highly effective in capturing the complex spatiotemporal patterns inherent in the MoCap data. 
Notably, many of these methods adopt a spatiotemporal transformer architecture, where attention is applied both across body joints within each frame (spatial) and across frames (temporal), enabling the model to simultaneously capture inter-joint relationships and motion dynamics. 
For example, PoseFormer \parencite{Zheng2021} pioneered the use of a pure spatiotemporal transformer architecture for 3D human pose estimation, MixSTE \parencite{Zhang2022MixSTE} alternates temporal and spatial attention in a seq2seq design, and STCFormer \parencite{Tang2023} further decomposes spatiotemporal modeling into parallel pathways. In summary, these works highlight the effectiveness of spatiotemporal transformers for human motion modeling.
\section{Problem Description and Solution}
\label{sec:problem}

\begin{figure*}[t]         
  \centering              
  \includegraphics[width=0.6\textwidth]{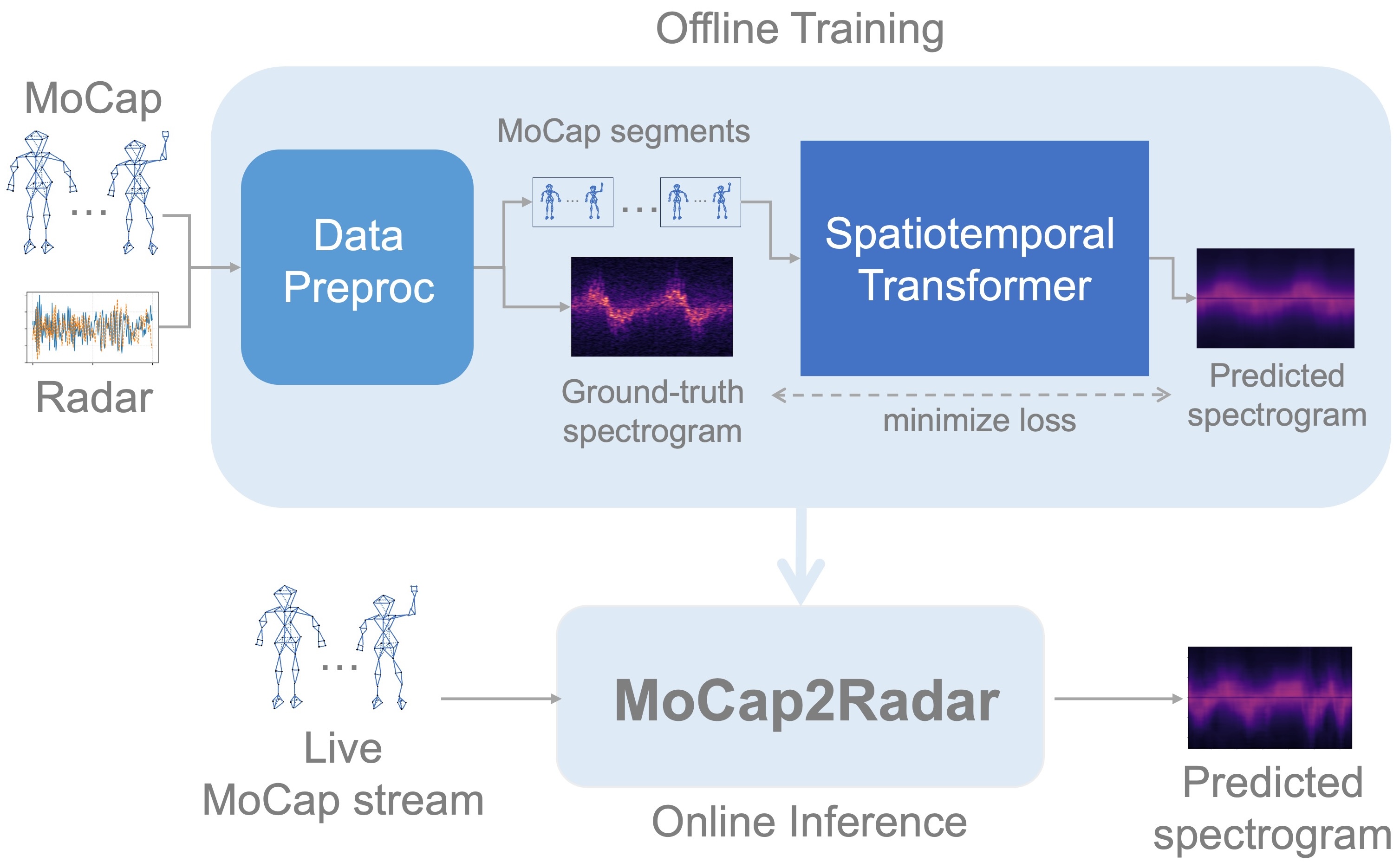}
  \caption{Overview of MoCap2Radar pipeline.}
  \label{fig:overview}
\end{figure*}

We cast the MoCap‑to‑radar translation as a sequence‑to‑sequence regression problem, where the objective is to infer the corresponding radar spectrogram, given a sequence of MoCap data. Formally, the objective is to learn the mapping:
\begin{align*}
  f_\theta:\;
  \bigl\{\boldsymbol{\mathcal X}_t \in \mathbb R^{W\times M\times D}\bigr\}_{t=1}^{T}
  \longrightarrow   
  \bigl\{{\mathbf{s}}^{\ast}_t \in \mathbb R^{F}\bigr\}_{t=1}^{T},
\end{align*} 
where $\boldsymbol{\mathcal X}_{t}$ denotes the $t$-th MoCap sequence of length $W$ frames;  each frame contains $M$ markers, and each marker is represented by a \(D\)-dimensional feature vector and $\mathbf s^{\ast}_{t}$ is the target doppler spectrum associated with the same time window, expressed as a real-valued vector of $F$ frequency bins. Note that a window of length  $W$ samples yields exactly $W$ frequency bins in the two-sided Short-Time Fourier Transform (STFT; hence, $F=W$. The following subsections detail how these input–output pairs are constructed and how the network \(f_{\theta}\) is trained.

Let a raw continuous MoCap stream be represented as $\mathbf{X} \in \mathbb{R}^{N \times M \times D}$, where $N$ is the number of frames and $M$ \& $D$ are defined above. The corresponding radar signal is represented as  $\mathbf{Y} \in \mathbb{C}^{N}$, i.e., each entry is a complex-valued sample.  

Figure~\ref{fig:overview} delineates the proposed machine learning pipeline of our MoCap2Radar solution, which unfolds in two phases: An offline training phase that, once the model has converged, is followed by an online inference phase. Both phases leverage two common modules: a data preprocessing module that prepares the data to be input to the model, and a spatiotemporal transformer that performs the MoCap-to-radar translation. We describe these two modules in the next two subsections.

\subsection{Data Preprocessing}
The data preprocessing block aligns the MoCap and radar streams, $\mathbf{X}$ and $\mathbf{Y}$ respectively, resamples and windows them with shared parameters, and produces temporally aligned snippet pairs that serve as the model’s supervised training examples.

First, the MoCap stream $\mathbf{X}$ is up-sampled from its native sampling frequency $F_m \! = \! 250~\text{Hz}$ to the radar rate $F_r \! = \! 256~\text{Hz}$ via linear interpolation, yielding the resampled sequence $\widetilde{\mathbf{X}}$ that now shares a common temporal grid with the radar signal. Next, we fix the STFT parameters for the radar signal—a window of length $W$ and hop $H$—and then segment the MoCap stream with the same window length and hop to ensure temporal alignment. Sliding a window of length $W$ over the sequence $\widetilde{\mathbf{X}}$ with hop $H$ yields
\begin{align*}
    \boldsymbol{\mathcal X} &\in \mathbb R^{T\times W\times M\times D}, \quad
    T = \Bigl\lfloor \frac{N-W}{H} \Bigr\rfloor + 1 \ \ .
\end{align*}

For each such MoCap snippet, the corresponding complex radar I/Q samples \(\mathbf Y\) are transformed by an STFT with the same $W$ and $H$.  The STFT is returned in its two‑sided form and scaled to spectral density, yielding the spectrogram with $F=W$ uniformly spaced frequency bins. To facilitate visualization of doppler shifts, we circularly shift the frequency axis so that the zero‑frequency bin is centered, producing $\mathbf{S}_c \in \mathbb{C}^{F\times T}$ with negative and positive frequencies symmetrically arranged about the origin.  The complex spectrum is then converted to an energy representation by taking its absolute value and a square‑root compression is applied to reduce its dynamic range. The resulting spectrogram, denoted by $ \mathbf{S} = \sqrt{\lvert \mathbf S_c\rvert} \in \mathbb{R}^{F\times T}$, highlights low‑energy components while preserving the relative intensity of dominant doppler features, making it well suited for both qualitative inspection and subsequent learning tasks. Transposing yields ${\mathbf{S}}^{\ast}=\mathbf{S}^{\top} \in \mathbb{R}^{T \times F}$ so that it is temporally aligned with the corresponding MoCap snippet.

\begin{figure*}[t!]         
  \centering              
  \includegraphics[width=\textwidth]{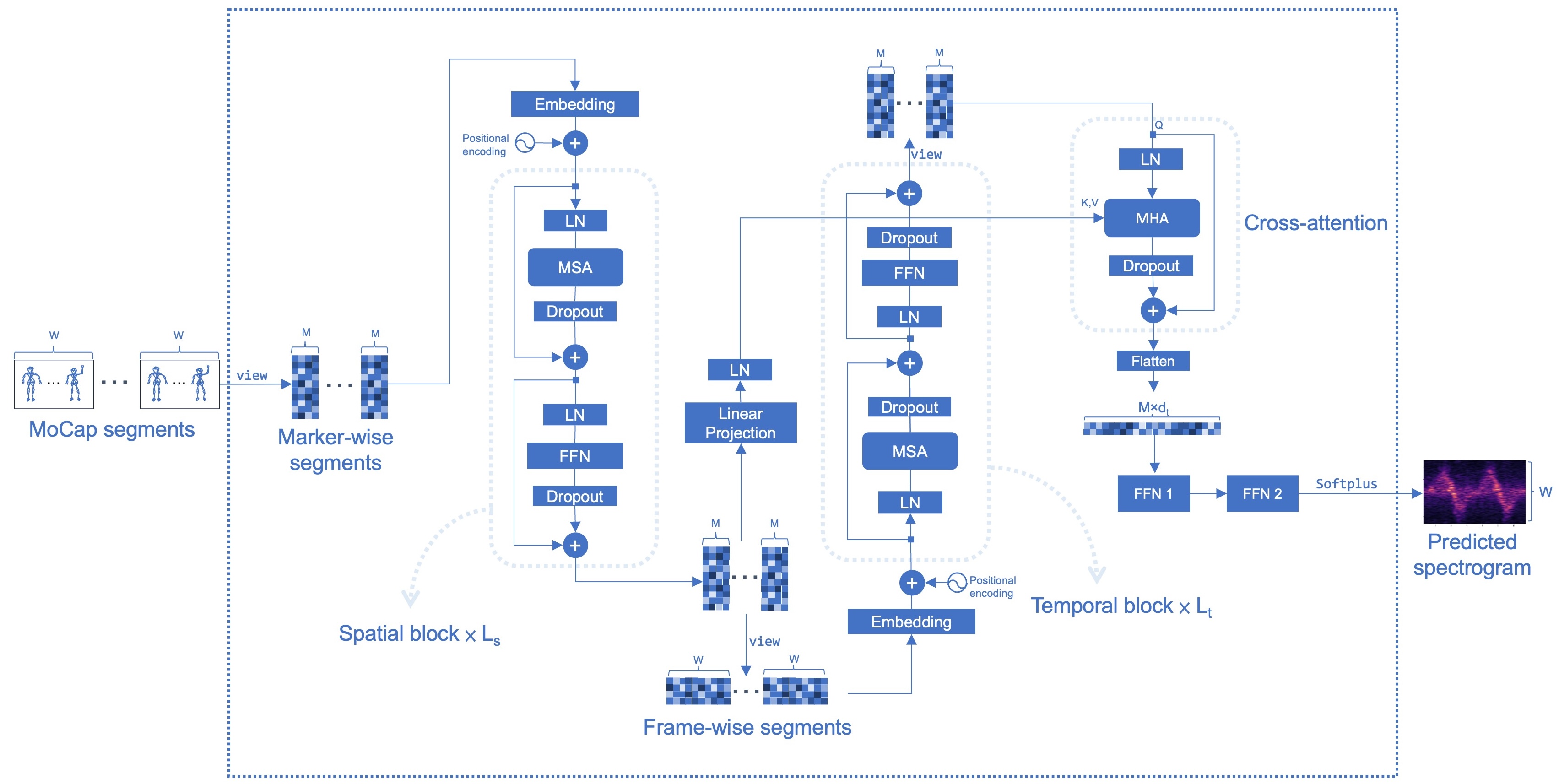}
  \caption{Spatiotemporal transformer architecture.}
  \label{fig:model}
\end{figure*}

\subsection{Spatiotemporal Transformer}

We propose a spatiotemporal architecture for our transformer model. 
It begins with a stack of $L_s$ spatial transformer encoder blocks, which learn frame-wise, joint-to-joint relationships. 
The resulting representations are then passed to a stack of $L_t$ temporal transformer decoder blocks, which capture long-range motion dynamics across the sequence. 
Finally, a lightweight multilayer perceptron converts the aggregated spatiotemporal features into a doppler-spectrum prediction. 
The two transformer blocks mirror the original transformer design \parencite{Vaswani2017}, except that we adopt the pre-norm variant, which places layer normalization before the attention and feed-forward blocks, a modification that improves gradient flow and training stability and is now common practice \parencite{Xiong2020}.

For notational simplicity, in the following detail descriptions of the network components, we omit the dimension $T$.

\vspace*{-1.5mm}
\subsubsection{Spatial Block}

As illustrated in Figure~\ref{fig:model}, the spatial block processes the $M$ markers in a single frame with a stack of $L_s$ layers.  Therefore, we fix  the frame index $w \in\{1, \ldots, W\}$ and collect the marker-wise segment $\mathbf{M}_w:=\mathcal{X}[w,:,:] \in \mathbb{R}^{M\times D}$,  whose $m$-th row contains the $D$-dimensional feature vector of marker $m$ at frame $w$. 

\ourPara{Marker embedding and positional encoding:} Each marker is linearly projected into $d_s$ dimensions and the resulting vectors are subsequently augmented with a positional encoding. Let $W_{s}^{(0)}\!\in\!\mathbb R^{D\times d_s}$ and
$b_{s}^{(0)}\!\in\!\mathbb R^{d_s}$ be learnable parameters. Then,
\[
  \widetilde{\mathbf M}_w^{(0)}
  = \mathbf M_{w} W_{s}^{(0)}
    + \mathds 1_M\, b_{s}^{(0)\!\top} + \mathbf{E}_{\mathrm{m}}
  \;\in\;\mathbb R^{M\times d_s},
\]
where $\mathds 1_M$ denotes an $M$‑dimensional column vector of ones, $ \mathbf{E}_{\mathrm{m}} \in \mathbb R^{M\times d_s}$ is the marker-wise positional encoding.

\ourPara{Pre-norm transformer layers:} The subsequent computations follow the standard pre-norm transformer block, comprising layer normalization ($\operatorname{LN}$) \parencite{ba2016layer}, multi-head self-attention ($\operatorname{MSA}$), residual dropout \parencite{srivastava2014dropout}, and a position-wise feed-forward network ($\operatorname{FFN}$). Each spatial transformer layer, $\ell_s = 1,\dots,L_s$, computes:
\vspace*{-1.5mm}
\begin{equation*}
\begin{aligned}
  \mathbf U_w^{(\ell_s)} &= \operatorname{LN}\bigl(\widetilde{\mathbf M}^{(\ell_s-1)}_w\bigr),\\
  \mathbf V_w^{(\ell_s)} &= \operatorname{MSA}\bigl(\mathbf U^{(\ell_s)}_w\bigr),\\
   \widehat{\mathbf M}_w^{(\ell_s)} &= \widetilde{\mathbf M}^{(\ell_s-1)}_w + \operatorname{Dropout}\bigl(\mathbf V^{(\ell_s)}_w\bigr),\\
  \mathbf R_w^{(\ell_s)} &= \operatorname{LN}\bigl( \widehat{\mathbf M}^{(\ell_s)}_w\bigr),\\
   \mathbf F_w^{(\ell_s)} &= \operatorname{FFN}\bigl(\mathbf R^{(\ell_s)}_w\bigr),\\
  \widetilde{\mathbf M}_w^{(\ell_s)} &=   \widehat{\mathbf M}_w^{(\ell_s)} + \operatorname{Dropout}\bigl(\mathbf F^{(\ell_s)}_w\bigr).
\end{aligned}
\end{equation*}
In $\operatorname{MSA}$ we use $H_s$ heads, each of width $d_h \! = \! d_s/H_s$.
For each head $h \in H_s$, the query, key, and value matrices are:
\vspace*{-1.5mm}
\begin{equation*}
\begin{aligned}
\mathbf Q_{w,h}^{(\ell_s)}=\mathbf U_w^{(\ell_s)}W_h^{(\ell_s)Q}, \\
\mathbf K_{w,h}^{(\ell_s)}=\mathbf U_w^{(\ell_s)}W_h^{(\ell_s)K},  \\
\mathbf V_{w,h}^{(\ell_s)}=\mathbf U_w^{(\ell_s)}W_h^{(\ell_s)V}, 
\end{aligned}
\end{equation*}
where $W_h^{\left(\ell_s\right) Q}, W_h^{\left(\ell_s\right) K}, W_h^{\left(\ell_s\right) V} \in \mathbb{R}^{d_s \times d_h}$, 
and the head output is
$$  \mathbf O_{w,h} ^{(\ell_s)}=
  \operatorname{Softmax}\bigl(\mathbf Q_{w,h}^{(\ell_s)}\mathbf K_{w,h}^{(\ell_s)\!\top}/\sqrt{d_h}\bigr)\mathbf V_{w,h}^{(\ell_s)} .
$$
The concatenated matrix
$[\mathbf O_{w,1}^{(\ell_s)} \|\dots \|\mathbf O_{w,H_s}^{(\ell_s)}]$
is then projected by
$W^{(\ell_s)O}\in\mathbb R^{H_s d_h\times d_s}$
to yield the block output
$\mathbf V_w^{(\ell_s)}=[\mathbf O_{w,1}^{(\ell_s)} \|\dots\|\mathbf O^{(\ell_s)}_{w, H_s}]W^{(\ell_s)O}$.
The position-wise FFN has a hidden width $d_s$ and ReLU activation.
The FFN applies the same two-layer MLP to every marker, computing $F_w^{(\ell_s)}$ as 
\[
  \operatorname{ReLU}\bigl(
      \mathbf R_w^{(\ell_s)} W_{1}^{(\ell_s)}
      + \mathds 1_M b_{1}^{(\ell_s)\!\top}
  \bigr)\, W_{2}^{(\ell_s)}
  + \mathds 1_M b_{2}^{(\ell_s)\!\top},
\]
where $W_{1}^{(\ell_s)},\,W_{2}^{(\ell_s)} \in \mathbb R^{d_s\times d_s}$ and $b_{1}^{(\ell_s)},\,b_{2}^{(\ell_s)} \in \mathbb R^{d_s}$.
After the last spatial layer $\left(\ell_s=L_s\right)$ we obtain, for each frame $w$, a marker-level feature matrix that is defined by
\[
  \mathbf{A}_w:=\widetilde{\mathbf M}_w^{\left(L_s\right)} \in \mathbb{R}^{M \times d_s}.
\]
Concatenating these matrices along the frame produces 
\[
  \mathbf{A}=\left[\mathbf{A}_1, \ldots, \mathbf{A}_W\right]^{\top} \in \mathbb{R}^{W \times M \times d_s},
\] 
which serves as the input to the subsequent temporal block.

\subsubsection{Temporal Block}

In this block, the model decodes the spatial features from the previous block and learns the motion dynamics across the $W$ frames with a stack of $L_t$ layers. Hence, we fix a marker index \(m\in\{1,\dots,M\}\).
Collecting its slice
\(\mathbf A_{[:,m,:]}\) produces the frame-wise segments $\mathbf{T}_m \in \mathbb{R}^{W \times d_s}$ whose $w$-th row is the $d_s$-dimensional feature vector of marker $m$ produced by the spatial block at frame $w$.

\ourPara{Marker embedding and positional encoding:} Each frame is linearly projected to a $d_t$-dimensional embedding, after which a positional encoding:
$$\widetilde{\mathbf{T}}_m^{(0)}=\mathbf{T}_m W_t^{(0)}+\mathds{1}_W b_t^{(0) \top}+\mathbf{E}_{\mathrm{f}} \in \mathbb{R}^{W \times d_t}~~,$$ where
$W_{t}^{(0)}\!\in\!\mathbb R^{d_s\times d_t}$, $b_{t}^{(0)}\!\in\!\mathbb R^{d_t}$ refers to learnable parameters, $\mathds{1}_W$ denotes an $W$-dimensional column vector of ones, $\mathbf{E}_{\mathrm{f}} \in \mathbb{R}^{W\times d_t}$ is the frame-wise positional encoding.

\ourPara{Pre-norm transformer layers:}
The temporal block uses the same sequence of operations as the spatial block; however, all weights are re-initialized and trained independently, so that temporal attention learns frame-to-frame relations while sharing no parameters with the spatial block. 
For each temporal transformer layer $\ell_t = 1,\dots,L_t $:
\vspace*{-1.5mm}
\begin{equation*}
\begin{aligned}
& \mathbf{U}_m^{(\ell_t)}=\operatorname{LN}\bigl(\widetilde{\mathbf{T}}_m^{(\ell_t-1)}\bigr), \\
& \mathbf{V}_m^{(\ell_t)}=\operatorname{MSA}\bigl(\mathbf{U}_m^{(\ell_t)}\bigr), \\
& \widehat{\mathbf{T}}_m^{(\ell_t)}=\widetilde{\mathbf{T}}_m^{(\ell_t-1)}+\operatorname{Dropout}\bigl(\mathbf{V}_m^{(\ell_t)}\bigr), \\
& \mathbf{R}_m^{(\ell_t)}=\operatorname{LN}\bigl(\widehat{\mathbf{T}}_m^{(\ell_t)}\bigr), \\
& \mathbf{F}_m^{(\ell_t)}=\operatorname{FFN}\bigl(\mathbf{R}_m^{(\ell_t)}\bigr), \\
& \widetilde{\mathbf{T}}_m^{(\ell_t)}=\widehat{\mathbf{T}}_m^{(\ell_t)}+\operatorname{Dropout}\bigl(\mathbf{F}_m^{(\ell_t)}\bigr).
\end{aligned}
\end{equation*}
After $\ell_t =L_t$, the result is defined by 
\[
  \mathbf{B}_m:=\widetilde{\mathbf T}_m^{\left(L_t\right)} \in \mathbb{R}^{W \times d_t}.
\]
Concatenating these matrices along the frame produces 
\[
  \mathbf{B}=\left[\mathbf{B}_1, \ldots, \mathbf{B}_M\right]^{\top} \in \mathbb{R}^{M \times W \times d_t},
\]
which processes into the next block.

\begin{figure*}[t!]
  \centering
  \begin{subfigure}[t]{0.32\textwidth}
    \centering
    \adjustbox{raise=0.3cm}{%
      \includegraphics[width=\linewidth]{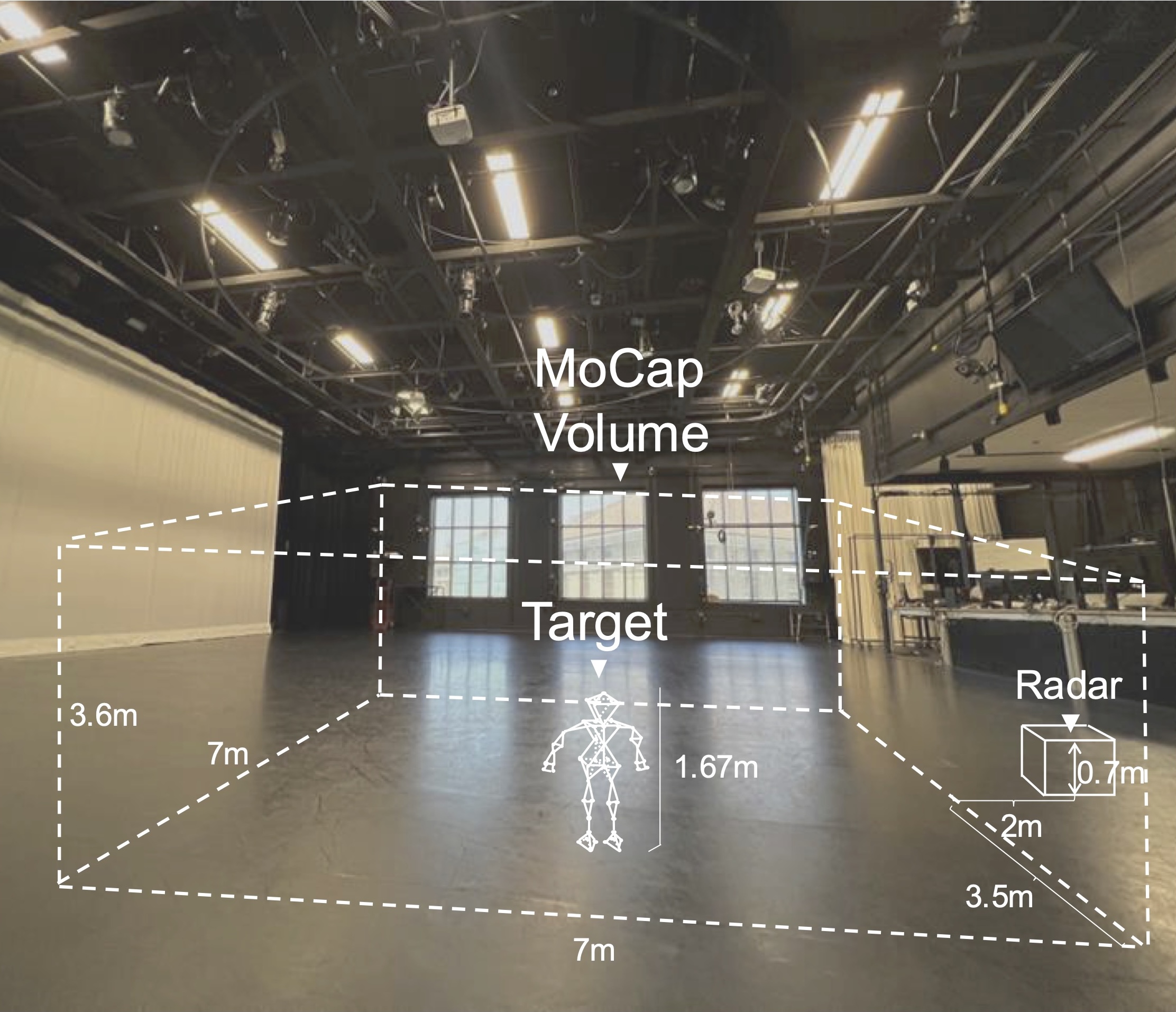}
    }
    \caption{MoCap and radar lab}
    \label{fig:setup-photo}
  \end{subfigure}
  \begin{subfigure}[t]{0.32\textwidth}%
\hspace{0.3cm}
    \centering
    \includegraphics[width=.8\linewidth]{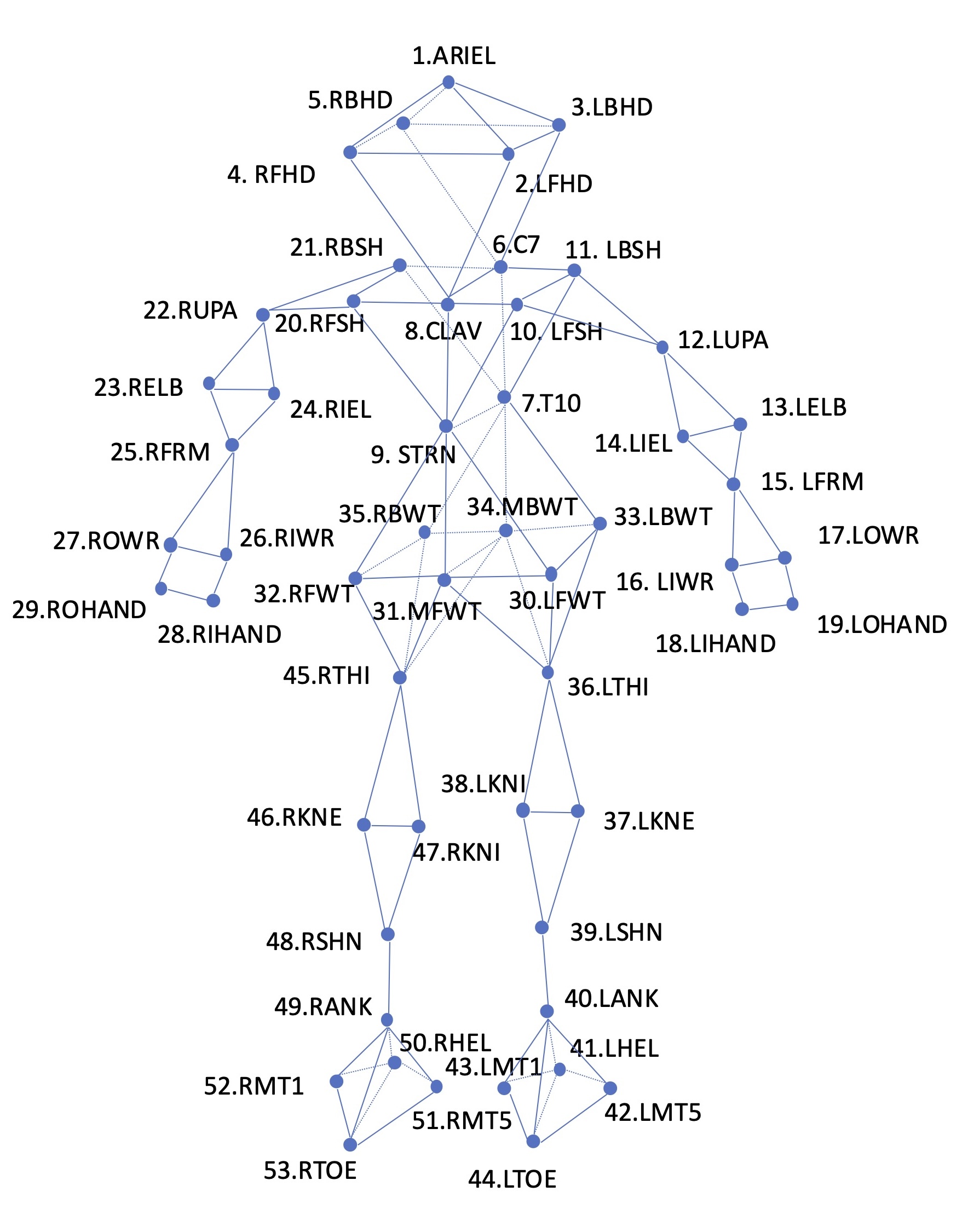}
    \caption{Vicon MoCap skeleton (53 markers)}
    \label{fig:setup-skeleton}
  \end{subfigure}
  \hfill
  \begin{subfigure}[t]{0.32\textwidth}
    \centering
    \adjustbox{raise=0.7cm}{%
    \includegraphics[width=.8\linewidth]{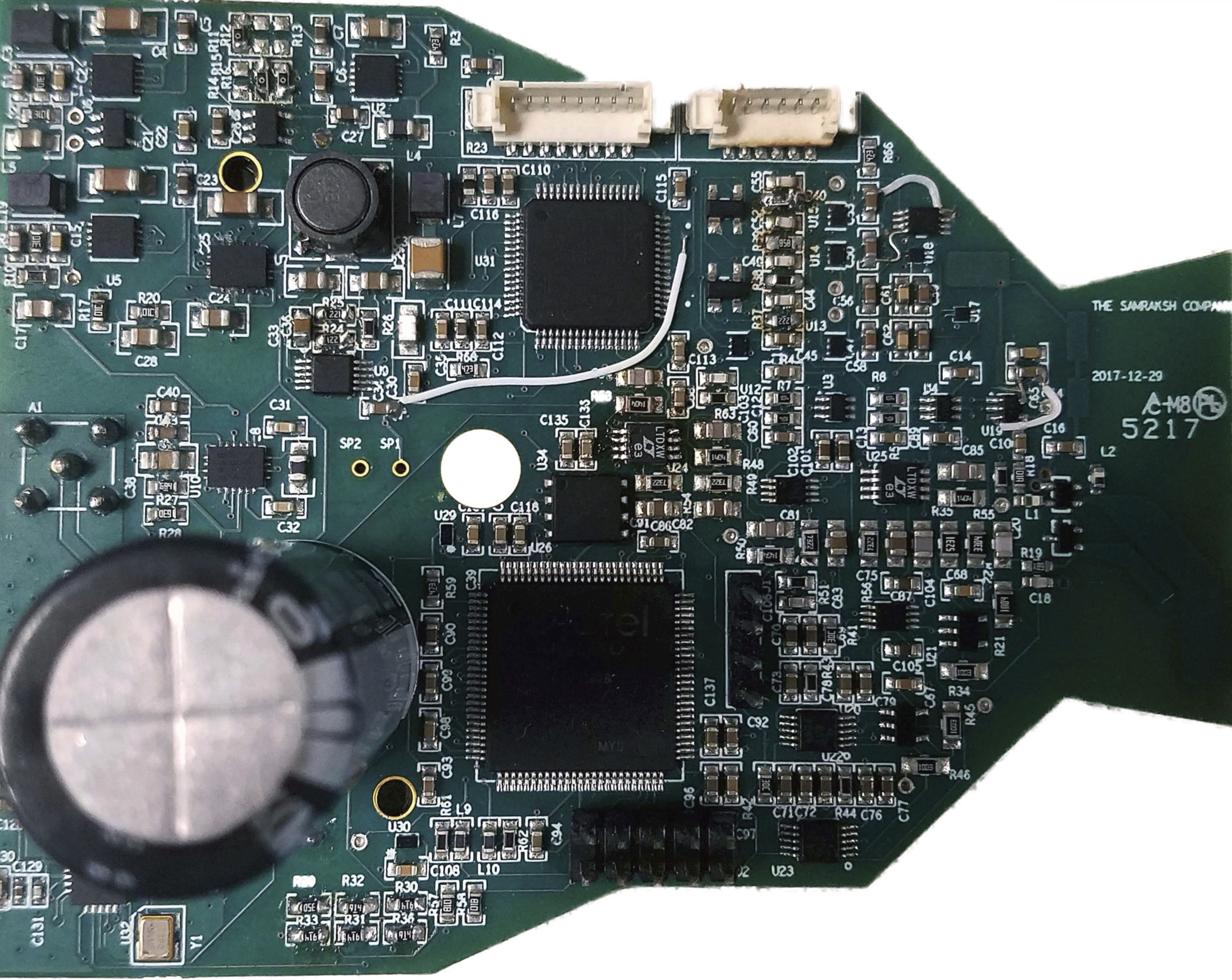}
    }
    \caption{Austere micro-doppler radar}
    \label{fig:setup-radar}
  \end{subfigure}
  \hfill
  \caption{Experimental setup for data collection.  
           (a) photo of MoCap and Radar Lab,   
           (b) marker layout on target, and
           (c) image of Austere radar.
           }
  \label{fig:setup}
\end{figure*}

\subsubsection{Cross‑Attention Fusion}

To integrate the complementary information learned by the spatial and temporal blocks, we apply a marker‑wise cross‑attention at each frame. From the spatial block, we have obtained each marker as \(\mathbf A_{w}\in\mathbb R^{M\times d_s}\). These spatial features are first projected to the temporal width and then layer-normalized:
\[
   \widetilde{\mathbf A}_{w}
   =\operatorname{LN}\!\bigl(\mathbf A_{w} W_{kv}+\mathds 1_M b_{kv}^{\!\top}\bigr),
\]
where
\[   
  w_{kv}\in\mathbb R^{d_s\times d_t},\;   b_{kv}\in\mathbb R^{d_t},
\]
and then used as keys and values,
\(\mathbf K_{w}=\mathbf V_{w}=\widetilde{\mathbf A}_{w}\).

We take the slice of temporal block output 
$\mathbf{R}_w :=\mathbf{B}[:,w,:] \in \mathbb{R}^{M \times d_t},$
whose $m$-th row is the $d_t$-dimensional temporal embedding of marker $m$ produced by the temporal transformer at that frame. The collection $\left\{\mathbf{R}_w\right\}_{w=1}^W$ is the queries used in the cross-attention fusion, i.e., $\mathbf{Q}_w = \mathbf{R}_w$. 

\ourPara{Multi-head cross-attention:}
Next, we compute the multi-head attention (MHA):
\begin{equation*}
    \begin{aligned}
   \mathbf C_{w} = \operatorname{MHA}(\mathbf Q_{w},\mathbf K_{w},\mathbf V_{w})
  \in\mathbb R^{M\times d_t}.       
    \end{aligned}
\end{equation*}
Specifically, using \(H_c\) heads of width \(d_h=d_t/H_c\), we form per-head projections $\mathbf Q_{w,h}= \mathbf Q_{w} W_{h}^{Q},$ $\mathbf K_{w,h}= \mathbf K_{w} W_{h}^{K},$ $\mathbf V_{w,h}= \mathbf V_{w} W_{h}^{V},$ where $W_{h}^{Q},W_{h}^{K},W_{h}^{V}\in\mathbb R^{d_t\times d_h}.$
The scaled dot-product attention for head $h \in H_c$ is
\[
  \mathbf O_{w,h}= \operatorname{Softmax}\bigl(
      \mathbf Q_{w,h}\mathbf K_{w,h}^{\!\top}/\sqrt{d_h}
    \bigr)\mathbf V_{w,h}\in\mathbb R^{M\times d_h}.
\]
Concatenating all heads and projecting back to width \(d_t\) yields
\[
  \mathbf C_{w}= 
  \bigl[\,\mathbf O_{w,1}\|\dots\|\mathbf O_{w,H_c}\bigr]W^{O},
\] 
where 
$$
  W^{O}\in\mathbb R^{H_c d_h\times d_t},  \mathbf C_{w}\in\mathbb R^{M\times d_t}.
$$
A residual connection followed by dropout and layer normalization yields the fused marker representation $$  \widehat{\mathbf C}_{w}
  = \operatorname{LN}\bigl(
      \mathbf R_{w} + \operatorname{Dropout}(\mathbf C_{w})
    \bigr)
  \in\mathbb R^{M\times d_t}.$$
Stacking the $W$ fused frames along the first mode yields  $$  \widehat{\mathbf C}=
  [\widehat{\mathbf C}_{1},\dots,\widehat{\mathbf C}_{W}]^{\top}
  \in\mathbb R^{W\times M\times d_t},$$
which is forwarded to the final inference block. 

\ourPara{Multilayer perceptron block:} The fused representation
\(
  \widehat{\mathbf C}\in\mathbb R^{W\times M\times d_t}
\)
is flattened across the marker–feature modes,
\[
  \mathbf C^{\mathrm{flat}}
    =\operatorname{reshape}\bigl(\widehat{\mathbf C}\bigr)
    \in\mathbb R^{W\times (M d_t)} .
\]
Then, a fully connected ReLU layer reduces the channel width to \(d_f\), computing $\mathbf Z$ as
\[
    \operatorname{ReLU}(\mathbf C^{\mathrm{flat}} W_{1}+b_{1})
  \in\mathbb R^{W\times d_f},
  \;
  W_{1}\in\mathbb R^{M d_t\times d_f}.
\]

A second fully connected layer serves as the output layer collapses the feature dimension ($d_{out}=1$),
\[
  \widehat{\mathbf s}
  =\mathbf Z W_{2}+b_{2}\in\mathbb R^{W},
  \;
  W_{2}\in\mathbb R^{d_f\times d_{out}}.
  \vspace*{-2mm}
  \]
  
Because the regression target is the square root of a magnitude spectrum, a $\operatorname{Softplus}$ activation ensures a non-negative output,
\[
  \widehat{\mathbf s}^{(+)}    =\operatorname{Softplus}\bigl(\widehat{\mathbf s}\bigr)
    \in\mathbb R^{W}.
  \vspace*{-2mm}
  \]
The resulting vector \(\widehat{\mathbf s}^{(+)}\) constitutes the predicted doppler spectrum for the entire window of \(W\) frames.

\vspace*{-0.1em}
\ourPara{Learning objective:}
Given the ground‑truth \(\mathbf s^{\ast}\in\mathbb R^{W}\), 
we minimize the mean‑squared error (MSE) loss:
  \vspace*{-3mm}
  \[
  \mathcal L_{\mathrm{MSE}}
    = \frac{1}{W}\sum_{w=1}^{W}
        \bigl(\widehat{{s}}^{(+)}_{w} - {{s}}^{\ast}
        _{w}\bigr)^{2}.
\]

\section{Experimental Data Collection}
\label{sec:data}

We conducted three real-world data collection sessions at the Advanced Computing Center for the Arts and Design (ACCAD), The Ohio State University, USA (Figure~\ref{fig:setup-photo}). We used its 12-camera Vicon MoCap system to record the kinematics of a single human target, while simultaneously acquiring radar signatures of the target using an Austere micro-doppler radar, made by The Samraksh Company (Figure~\ref{fig:setup-radar}) \parencite{roy2021one}. The Vicon system tracks $M=53$ reflective markers attached to the target’s body (Figure~\ref{fig:setup-skeleton}) at $F_m = 250$ frames per second, yielding time-stamped $(x,y,z)$ coordinates for each marker. The Austere is a homodyne radar designed for the $5.8$ GHz Industrial, Scientific, and Medical (ISM) band with $100$ MHz bandwidth, $12$~m omnidirectional range and a $100$~Hz doppler bandwidth.
It yields complex baseband returns—\emph{in-phase} $I$ and \emph{quadrature} $Q$ components—at $F_r = 256$ samples per second. The participant was an adult male, $167.6$ cm in height and $59$ kg in mass.
	
\ourPara{Session 1:}
The MoCap capture region is a rectangular box measuring $7\times 7\times 3.6\,\mathrm{m}$. The radar was placed outside this boundary at $(3.5 \mathrm{~m}, 2 \mathrm{~m})$ in the local coordinate frame and mounted $0.7 \mathrm{~m}$ above the floor with its antenna oriented toward the MoCap volume. The participant walked along each diagonal path, in both directions and at two speeds, the faster of which being approximately twice the nominal walking speed. This yields a total of 8 trial types. Each path–speed combination was repeated 10 times. 

\ourPara{Session 2:}
In this session, the participant again traversed the four diagonal paths, but each trajectory was repeated only twice.  The rest of the experimental setup remains unchanged. This also yielded a total of 8 trial types.

\ourPara{Session 3:}
The participant performed two free-walking trials and was instructed to vary speed arbitrarily throughout each recording (spontaneous accelerations/decelerations). 
Because the participant modulated speed freely during each trial, the two walking trials can be regarded as independent. 
All other experimental conditions remained unchanged. This yielded a total of $2$, relatively long, and independent trials.
	
Since precise temporal synchronization and alignment between the MoCap and radar streams is essential for supervised learning, a procedure was adopted to guarantee frame-level synchronization. 
Immediately after the system starts to record, the Vicon controller issues a pulse that is passed through a USB-C cable to the radar data collection system. 
The rising edge is logged in both devices, providing a common epoch for subsequent time stamps. 

\begin{table}[b!]
\centering
\caption{Model hyperparameters for all experiments.}\label{tab:hyperparams}
\setlength{\tabcolsep}{5pt}
  \begin{tabular}{lc}
    \hline\hline
    \textbf{Parameter} & \textbf{Value} \\
    \hline
    Spatial embed.\ dim.\ ($d_s$)   & 64   \\
    Temporal embed.\ dim.\ ($d_t$)  & 128  \\
    Feed-forward width ($d_f$)      & 256  \\
    Spatial heads ($H_s$)           & 2    \\
    Temporal heads ($H_t$)          & 4    \\
    Spatial layers ($L_s$)          & 2    \\
    Temporal layers ($L_t$)         & 4    \\
    Dropout $p$ (all layers)        & 0.3  \\
    Batch size                      & 8    \\
    Output dim.\ ($d_{\text{out}}$) & 1    \\
    \hline\hline
  \end{tabular}
\end{table}

All machine learning experiments (including the ablation studies) adopt the hyperparameters listed in Table~\ref{tab:hyperparams}.
Training is performed without early stopping for 50~epochs with the Adam optimizer (initial learning rate $\eta_{0}=3\times10^{-4}$; weight decay $=10^{-5}$). 
A \texttt{OneCycleLR} scheduler \parencite{smith2019super} is used: $\eta$ is increased linearly from $3\times10^{-7}$ to the base value $\eta_{0}$ during the first 10\,\% of updates and then decreased linearly to $3\times10^{-10}$ over the remaining 90\,\%. This strategy has been shown to achieve better performance compared to a fixed learning rate. The schedule spans the full 50~training epochs. All experiments are conducted on a single Google Colab instance equipped with an NVIDIA T4 GPU (16 GB), demonstrating the potential of both training and inference in edge computing environments.

\section{MoCap2Radar Performance Evaluation}
\label{sec:validation}

\begin{figure}[b!]                
  \centering
   \includegraphics[width=0.9\columnwidth]{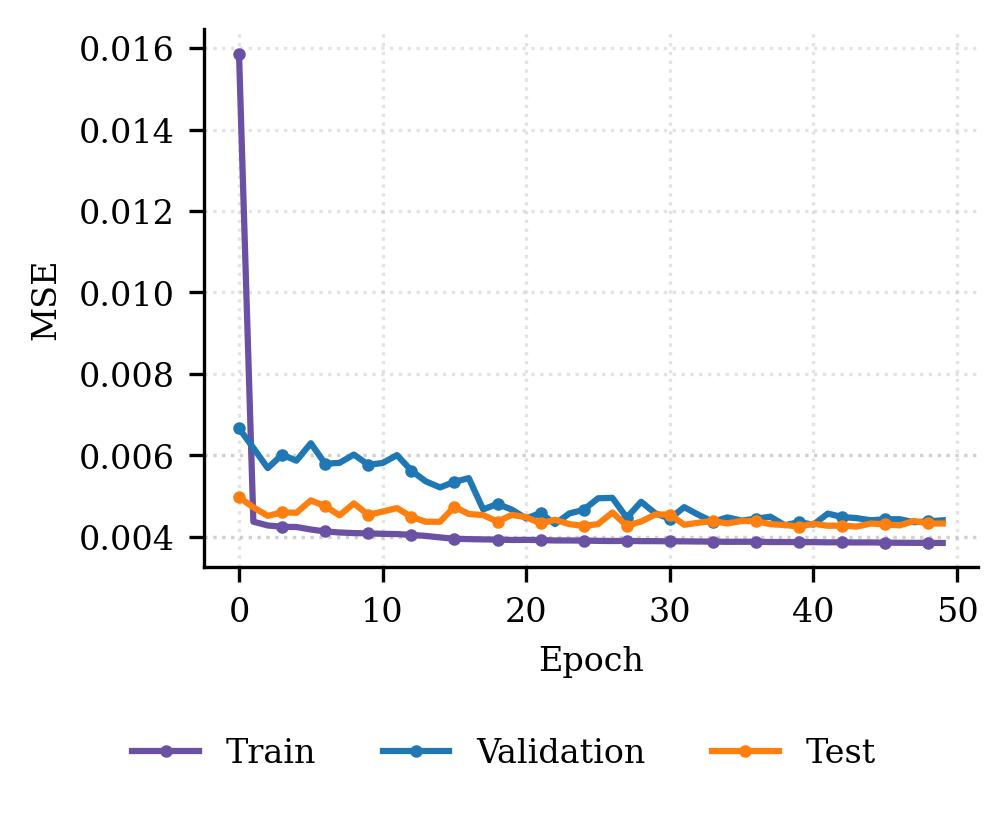}
   \caption{Loss curves for training, validation, and test sets over 50 epochs.}
   \label{fig:loss}
\end{figure}

We used data from all sessions for training and evaluation.
Session~1 was treated as the training set, with $10 \%$ of its samples randomly selected as a validation set.  
Session~2 was employed as the testing set; its recordings capture the same diagonal-walk motion as Session 1, so the data should be statistically similar but entirely separate from the training set. 
From Session~3, we select a randomly meandering walk as an indication of the model's generalization to the larger class of arbitrary walking.
We also examine the convergence behavior of the model, as well as inference performance, and conduct an ablation study to assess the necessity of the proposed architecture.
The following subsections present our empirical findings.

\begin{figure*}[th!]
    \centering
      \includegraphics[width=\textwidth]{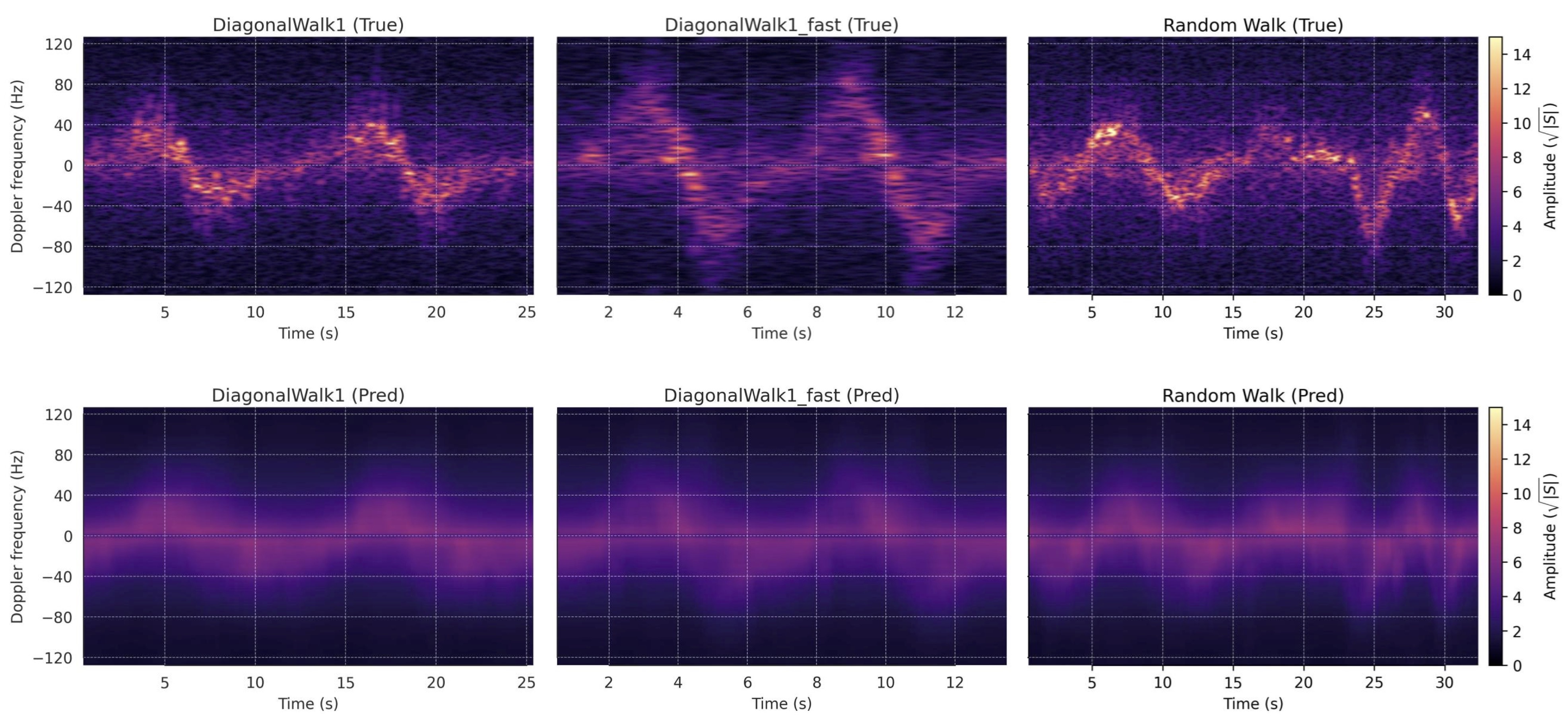}
      \caption{%
        \textit{Columns}—\textbf{Left}: diagonal‑walk at nominal speed
        (complete sequence);
        \textbf{Center}: diagonal‑walk at higher speed
        (complete sequence).
        \textbf{Right}: random‑walk sequence
        (randomly sampled contiguous frames);
        \textit{Rows}—top: measured radar spectrograms;
        bottom: spectrograms predicted by the proposed model.}

    \label{fig:result}
\end{figure*}

\subsection{Training, Validation and Testing}
As seen in Figure~\ref{fig:loss}, all three loss curves decrease steadily and plateau between epochs 20~and 30. The training, validation and test gap remains modest and stable, indicating that the model has converged and generalizes well with only limited over-fitting. 
Specifically, training loss collapses rapidly to $\approx 4.3 \times 10^{-3}$ within the very first few epochs and then forms a flat plateau, showing that learning has essentially converged and almost no further capacity is being fit. Validation loss follows the same trajectory but plateaus higher than the training floor; the gap stays instead of widening, which signals only mild over-fitting. 
Test loss tracks the validation curve almost identically after epoch $30$ and the ratio remains approximately constant and small, implying the validation split is representative of the unseen data and that minimal leakage is present.

\subsection{Testing Evaluation}

As shown in Figure~\ref{fig:result}, the MoCap2Radar model predicts peaks that align with those in the ground-truth, indicating the model's ability to learn the dominate periodicity associated with walking. 
In Figure~\ref{fig:result}, even finer micro-doppler structures in the radar signature are learned, such as the gap observed between 2~and 4 seconds. However, the high-frequency components present in the examples of faster walking are truncated. This suggests that the model may be biased toward generating smoother outputs, potentially due to limitations imposed by the MSE loss function. 
(In follow-on work, we plan to explore spectral-correlation or perceptual loss functions to better preserve high-frequency doppler components.)

\begin{figure*}[tbh]                
  \centering
   \includegraphics[width=0.8\textwidth]{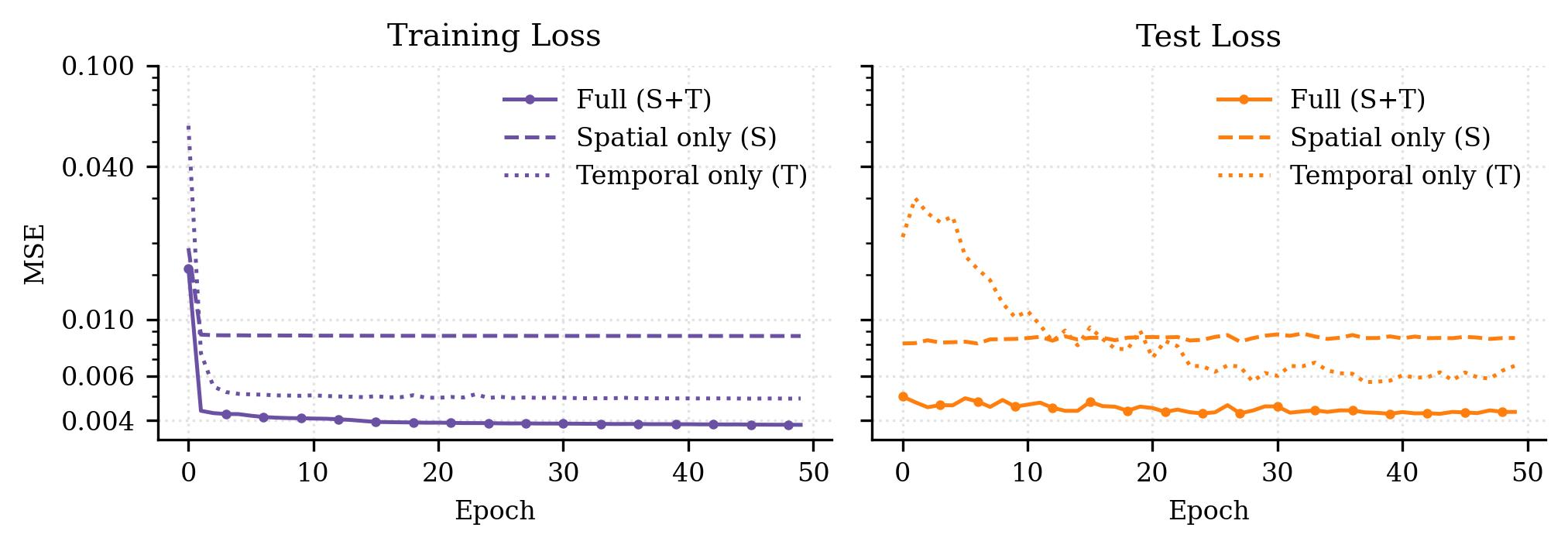}
   \caption{Loss curves over 50~epochs for the spatial‑only, temporal‑only, and spatiotemporal transformer models for the ablation study.}
   \label{fig:loss-train-test}
\end{figure*}

The results from the random walk case suggest that the model exhibits generalization.  
When analyzing the randomly meandering walks, we first observe that the motion, though oscillatory, is no longer periodic.
The three frequency peaks are not regularly spaced and differ significantly from those observed in the steady walks.
Remarkably, the model’s prediction is able to capture this variation in the dynamics, despite the fact that this style of walking was not included in the training set. 
This result strongly indicates that the model can generalize beyond the observed motions and successfully infer patterns it has not previously encountered.
Given the exploratory nature of this study and the small size of our dataset it is unlikely that generalization extends to dancing (or perhaps even to running), but it suggests that with a larger data set such generalization is likely.

\subsection{Ablation Studies}
To substantiate the effectiveness of the proposed spatiotemporal architecture, we performed a comprehensive ablation study that contrasted three model variants:  (i) Spatial‑only transformer (S), which leverages skeletal topology but discards temporal cues; (ii) Temporal‑only transformer (T), which exploits sequential dependencies while ignoring spatial structure; and (iii) the full (S+T) model, which jointly encodes spatial and temporal information. 
The training and test curves, depicted in Figure~\ref{fig:loss-train-test}, reveal three key observations.
First, the S+T model achieves the lowest MSE on both the training and test sets and converges substantially faster, underscoring its superior learning efficiency.
Second, removing temporal context, spatial‑only led to the highest MSE and the slowest convergence, indicating that spatial cues alone are insufficient for accurately modeling MoCap sequences.
Third, although the temporal‑only variant benefits from temporal continuity and attains a lower MSE than spatial‑only, its convergence is less stable and its final test error remains higher than that of the full model, reflecting limited generalization when spatial structure is omitted.
Collectively, these results provide empirical evidence that simultaneous modeling of spatial and temporal dependencies is essential for robust and generalizable MoCap-into-radar learning.

\section{Discussion and Concluding Remarks}
\label{sec:discussion}

\ourPara{The more challenging radar:}
The method presented here could be applied to a wide range of radars, but we have focused on a more challenging class of radars: low-complexity, single range-bin, doppler radars. 
These radars are inexpensive (i.e., \$10--\$100), tend to be low enough power for battery powered operations, and are almost exclusively used in edge sensing. 
They often provide superior doppler fidelity compared to traditional radars.
However, they output the superposition of the doppler shifts from all the moving objects within the field of view. 
As a result, although they provide sufficiently rich information to make subtle distinctions between complex multi-component motion patterns, disentangling movements sufficiently for downstream analysis is challenging.
Ironically, the simplicity of the hardware requires greater complexity in the software, but it also motivates the use of ML.

\ourPara{Difficulty in learning raw radar output:}
A key subtlety of this paper is in framing the problem as predicting the spectrogram of the radar signal rather than predicting the radar signal.
Learning a model to predict raw radar output is not practical with standard MoCap systems, because the phase of a return from an arbitrary reflection point is the remainder of twice the distance from that point to the radar, divided by the wavelength.
Phase differences of about 15\textdegree\ will typically materially alter the result.
The Austere radar was designed to work at 5.8~GHz, so its wavelength is 5.18~cm, implying a required location accuracy of about 1.1~mm.

Interestingly, the standard MoCap systems are slightly more accurate than this at the MoCap points.
However, since the signal is reflected from the entire body, this level of accuracy needs to be maintained for the estimates of all the points between the MoCap points.
Differences in human bodies, such as muscle shapes or the proportion of fat, cause differences in body shape that exceed this threshold, especially during motion.
In addition, the loss becomes insensitive to the geometric estimation if the phase error changes by more than 90\textdegree\ across the target, which breaks standard gradient descent training.
All of our attempts to train a model to directly estimate the raw radar output failed for these reasons.
At the very least, learning a model to directly predict the radar output would require many more MoCap points, probably a couple thousand.

A key breakthrough was realizing that predicting micro-doppler results did not have these problems and that it is difficult to construct an application using single range-bin micro-doppler radars that needs more than micro-doppler results.

\ourPara{The human body is part of our model:}
Our ablation experiments indicate that temporal analysis is necessary to achieve useful results.
This is expected since the spectrogram is fundamentally about rates of motion, and it is expected that the accuracy of inferring motion from a single moment in time is quite limited.
However, they also indicate that spatial analysis significantly improves the results.
This implies that the model is doing more than assigning radar cross sections (RCSes) to the MoCap points.
Our model is learning meaningful results based on the relative position of MoCap points.

We hope to complete and report on deeper interpretability results soon.
But it appears that the model is learning things such as that the orientation of the forearm can be inferred from the relative position of the wrist and the elbow and that this orientation significantly affects the RCS of that portion of the body.
The subsumption of a model of the human body is a key reason that it is not practical to deduce our model from first principles of physics and engineering.

\ourPara{Future work:} The fact that our model generalizes to motion patterns outside the domain of training suggests that this approach should allow for a kind of transfer learning where new applications are learned, or extended, from motion capture data without accompanying radar data.
This is useful because of the availability of comparatively large amounts of motion capture data covering diverse motion patterns (or because of the scarcity of relevant radar data).

While we do not expect our current model will generalize to all motion patterns, we do expect that in almost all applications the process presented here will lead to an equivalent model with far less data than would be required to learn the full application model.
This is because for interesting applications the proportion of the model devoted to understanding the range of motions in the environment will be large compared to the portion of the model required to convert that motion into estimations of the spectrogram.
As a result, learning the higher level portions of the model will require more data than learning the portions of the model corresponding to the MoCap2Radar model.
If our model is sufficiently generalizable much of the data for the higher level portions of the application model can be MoCap data (i.e., without corresponding radar data). 
We hope to demonstrate this more fully in follow on work.

\small
\printbibliography

\end{document}